\documentclass[11pt]{article}
\usepackage{acl}
\usepackage{times}
\usepackage{latexsym}
\usepackage[T1]{fontenc}
\usepackage[utf8]{inputenc}
\usepackage{microtype}
\usepackage{inconsolata}
\usepackage{booktabs}
\usepackage{amsmath,amssymb}
\usepackage{multirow}
\usepackage{graphicx}
\usepackage{enumitem}
\usepackage{dblfloatfix}
\usepackage{hyperref}

\usepackage[T1]{fontenc}
\usepackage{textcomp}
\usepackage{xcolor}
\usepackage{listings}
\usepackage[most]{tcolorbox}
\tcbuselibrary{listings,skins,breakable}
\usepackage{caption}
\usepackage{multicol}

\lstdefinestyle{promptcompact}{
  basicstyle=\ttfamily\footnotesize,
  breaklines=true,
  breakatwhitespace=true,
  columns=fullflexible,
  keepspaces=true,
  showstringspaces=false,
  tabsize=2,
  upquote=true,
  aboveskip=0pt,
  belowskip=0pt
}

\newtcblisting{promptbox}[2][]{
  enhanced,
  breakable,
  listing only,
  listing engine=listings,
  listing options={style=promptcompact},
  width=\columnwidth,
  colback=gray!4,
  colframe=gray!55!black,
  boxrule=0.7pt,
  arc=1.5mm,
  left=1.6mm,
  right=1.6mm,
  top=2.2mm,
  bottom=1.6mm,
  title={#2},
  fonttitle=\bfseries\rmfamily\small,
  coltitle=white,
  colbacktitle=gray!65!black,
  boxed title style={
    colback=gray!65!black,
    colframe=gray!65!black,
    arc=1mm,
    boxrule=0pt
  },
  attach boxed title to top left={
    xshift=3mm,
    yshift=-2mm
  },
  before skip=8pt,
  after skip=4pt,
  #1
}

\title{CARE-RL: Capability-Aware Reinforcement Learning for Mitigating Cross-Domain Conflicts}

\author{
Rui Zhang, Xinle Wu, Yao Lu \\
National University of Singapore \\
Singapore
}

\begin{document}
\maketitle

\begin{abstract}
Reinforcement learning (RL) with verifiable rewards has achieved strong progress in reasoning-oriented LLMs, but extending it to multi-domain RL remains challenging due to reward unreliability in non-verifiable tasks and capability interference across domains. We propose CARE-RL to combine protocol-aware reward generation with capability-aware optimization for mitigating cross-domain conflicts. For non-verifiable tasks, the Protocol-Aware Generative Reward Model (PA-GRM) constructs prompt-level evaluation protocols and schemas before producing trace-conditioned rewards, enabling task-adaptive yet comparable evaluation of open-ended responses. For multi-domain optimization, Direction-Aware Capability Subspace Projection (DACSP) extracts historical capability directions from previous RL stages and modulates later updates by amplifying aligned components, suppressing conflicting components, and preserving orthogonal updates. Experiments across math, chat, and instruction-following benchmarks show that CARE-RL consistently outperforms standard multi-domain RL baselines, achieving Total Avg scores of 47.9 and 50.7 on Qwen2.5-7B and Qwen3-4B, respectively.
\end{abstract}

\section{Introduction}

Reinforcement learning (RL) has become a central paradigm for post-training large language models (LLMs), especially in large reasoning models~\citep{guo2025deepseek,kimi2025k1.5}. 
A key driver of this progress is Reinforcement Learning with Verifiable Rewards (RLVR), where rewards are derived directly from task outcomes rather than from reward models~\citep{wen2025rlvr}. For example, mathematical responses can be checked against reference answers, and generated code can be evaluated by unit tests. Because such rewards are grounded in task correctness, they are less exposed to subjective preference noise and reward-model bias. 
This has made RLVR effective in verifier-rich domains such as mathematics and coding~\citep{shao2024deepseekmath}, while recent studies have begun to push RLVR beyond narrowly verifiable tasks toward broader, multi-domain settings with free-form or less structured answers~\citep{su2025crossing,yu2025rlpr,gunjal2025rubrics}.

However, extending RLVR to multi-domain learning remains fundamentally challenging, because many real-world tasks are open-ended then non-verifiable. To provide rewards for such tasks, recent work has explored generative reward models~\citep{zhang2025generative} and rubric-based evaluation protocols~\citep{dr_tulu,gunjal2025rubrics}, replacing direct scalar scoring with fine-grained evaluation. 
However, these methods do not fully solve the reward construction problem as even non-verifiable tasks may require different evaluation protocols. For example, a story-writing prompt is often better judged holistically by its coherence, style and overall engagement, while a travel-planning prompt requires checking whether the response satisfies concrete constraints such as budget, schedule and user preferences. Therefore, applying a single fixed judging protocol to heterogeneous open-ended prompts can still produce unstable or misaligned rewards, even when the judge model itself is strong.
Beyond reward construction, multi-domain RL also requires a single model to learn across diverse capabilities and data distributions. Existing multi-domain RL methods typically adopt either joint or sequential training, but joint training may suffer from conflicts among multiple objectives, while sequential training may cause later updates to degrade capabilities acquired in earlier stages~\citep{cheng2025revisiting,yang2026domains,zheng2025spurious}. Therefore, multi-domain RL requires not only a more flexible reward construction mechanism for non-verifiable tasks, but also an optimization mechanism that preserves accumulated capabilities while adapting to new domains.

To address these challenges, we propose \textbf{CARE-RL}, a reward-adaptive and capability-aware cascade learning framework for mitigating conflicts in cross-domain RL. On the reward side, we introduce the Protocol-Aware Generative Reward Model (\textbf{PA-GRM}) for non-verifiable tasks. Instead of assigning rewards under a fixed evaluation scheme, PA-GRM first constructs a prompt-level evaluation protocol and scoring schema, and then generates response-level scoring traces under the protocol. By routing prompts between holistic and rubric-based judging, PA-GRM produces rewards that better match the structure of heterogeneous open-ended tasks. On the optimization side, we introduce Direction-Aware Capability Subspace Projection (\textbf{DACSP}). The parameter change of each completed RL stage encodes the capability acquired in that stage, and its leading singular directions approximate capability-increasing directions. DACSP uses them as a low-rank reference for later updates: amplifying aligned components, suppressing opposing ones and leaving orthogonal ones unchanged.
In this way, CARE-RL integrates prompt-adaptive reward construction with update modulation across cascade stages, aiming to provide more suitable supervision for non-verifiable tasks while reducing update components that conflict with previously learned domains.

Our contributions are summarized as follows:
\begin{itemize}[itemsep=2pt, topsep=2pt, parsep=0pt, partopsep=0pt]
    \item We propose \textbf{CARE-RL}, a unified framework that simultaneously addresses two key challenges in multi-domain RL: reward unreliability on non-verifiable tasks and cross-domain capability degradation.
    \item PA-GRM adaptively constructs per-prompt evaluation protocols, alleviating the gap of fixed evaluation schemes in heterogeneous tasks and providing more reliable rewards; DACSP extracts capability subspaces from prior RL stages and applies direction-aware modulation to subsequent updates, preventing the erosion of acquired capabilities.
    \item CARE-RL achieves the highest average scores on \texttt{Qwen2.5-7B}/\texttt{Qwen3-4B}, outperforming the strongest baseline by +1.0/+0.9 points.
\end{itemize}


\begin{figure*}[t]
    \centering
    \includegraphics[width=\textwidth]{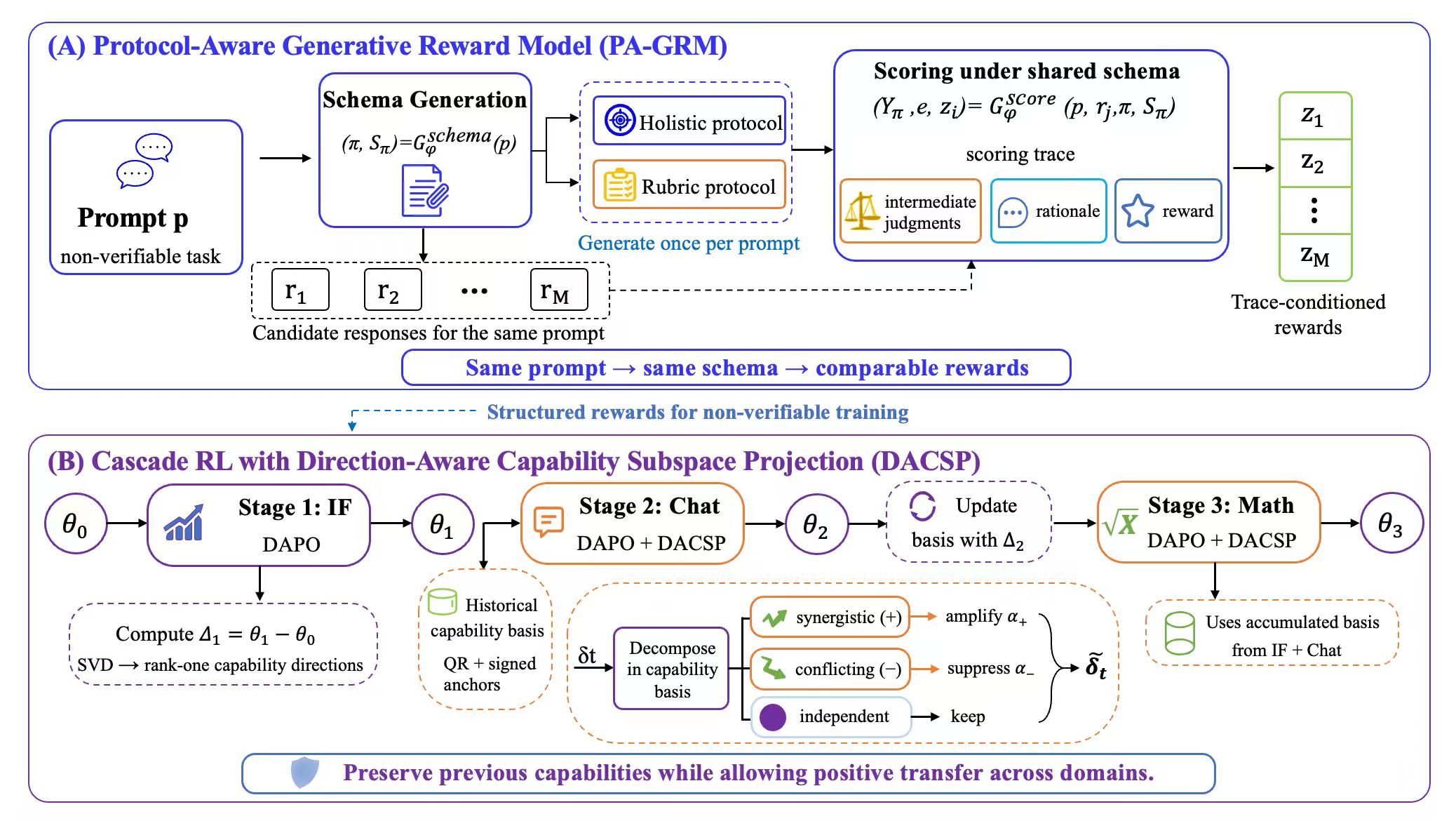}
    \caption{
    Overview of CARE-RL.
    (A) PA-GRM builds a prompt-level protocol and schema, then scores all candidates under the shared schema for trace-conditioned rewards.
    (B) Cascade RL over IF, Chat, and Math with DAPO; DACSP modulates later updates along historical capability directions, amplifying synergistic, suppressing conflicting, and preserving independent components.
    }
    \label{fig:overall_framework}
\end{figure*}

\section{Related Works}
\label{related works}

\paragraph{Reinforcement Learning towards General Domains.}
Recent LLM post-training research has extended RL from verifier-rich reasoning tasks to broader general-domain settings. In verifiable domains, pure RL and RLVR-style training have shown strong effectiveness for reasoning-oriented models~\citep{guo2025deepseek,shao2024deepseekmath}, while scalable RL algorithms such as DAPO and VAPO further improve advanced reasoning training~\citep{yu2026dapo,yue2025vapo}. Beyond narrow verifier-rich domains, recent systems broaden RL through cross-domain data construction and verifiability-aware data mixing~\citep{general_reasoner,nemotron_crossthink}. Other studies explore multi-domain RL settings that combine verifiable and non-verifiable training signals~\citep{zerogeneral2025,rlmt}. As the training scope expands, mixed-domain and sequential-domain optimization have become increasingly important, motivating methods based on gradient surgery, task balancing, and cascaded RL~\citep{mgs,mtgrpo,nemotron_cascade,nemotron_cascade2}. This reflects the shift from verifier-rich RL in single domain toward multi-domain RL over heterogeneous capabilities and reward sources.

\paragraph{Judge for Open-ended Problems.}
Open-ended reward modeling has also moved from direct scalar judging toward more structured evaluation. Reasoning-based reward models formulate judgment as a reasoning process, and branch-and-rethink mechanisms further improve difficult evaluations~\citep{rmr1,think_twice_brrm}. Rubric-based reward methods provide another structured approach by decomposing open-ended evaluation into explicit criteria; recent work shows that rubric feedback can support RL beyond verifiable domains and can be refined or evolved for long-horizon open-ended training~\citep{gunjal2025rubrics,dr_tulu,shen2026rrd}. Overall, these works indicate a trend from black-box scalar reward prediction toward more interpretable and structured evaluation processes.

\section{Methodology}
\label{Methodology}

As shown in Figure~\ref{fig:overall_framework}, CARE-RL organizes multi-domain RL as a cascade over multiple training domains. For verifiable domains, rewards are obtained from task-specific verifiers, while CARE-RL invokes PA-GRM to construct rewards for non-verifiable domains. After each RL stage, CARE-RL records the parameter change induced by that stage and uses DACSP to extract historical capability directions from the update. In later stages, before applying each optimizer step, DACSP decomposes the proposed update with respect to these historical directions and modulates its components according to their alignment. Therefore, PA-GRM determines how rewards are produced for open-ended tasks, while DACSP determines how updates are applied across stages. The remainder of this section first describes PA-GRM, then presents DACSP.

\subsection{Protocol-Aware Generative Reward Model}
\label{sec:protocol_aware_grm}

In order to route heterogeneous non-verifiable prompts to evaluation protocols that match their task structure, we introduce the PA-GRM, denoted as \(G_{\psi}\). PA-GRM decomposes reward generation into prompt-level protocol construction and response-level scoring. Given a prompt, it first selects an evaluation protocol and constructs the corresponding scoring schema; given a candidate response, it then generates a scoring trace under this prompt-specific schema, from which the final scalar reward is extracted. Thus, the evaluation standard is adapted to each prompt, while all candidate responses to the same prompt are judged under a shared schema. The complete prompt templates are provided in Appendix \ref{app:pagrm_prompts}.

We consider two evaluation protocols:
\[
\pi \in \{\textsc{holistic}, \textsc{rubric}\}.
\]
The protocol specifies how quality factors are operationalized during scoring. The \(\textsc{holistic}\) protocol is used for open-ended prompts that are difficult to decompose into independently checkable criteria, such as creative writing, where quality is better judged holistically. The \(\textsc{rubric}\) protocol is used for prompts with explicit structural or content requirements, where each requirement can be checked more independently.

Given a prompt \(p\), PA-GRM first determines the protocol and schema:
\[
(\pi,\mathcal{S}_{\pi})
=
G_{\psi}^{\mathrm{schema}}(p),
\]
where \(\mathcal{S}_{\pi}\) denotes the evaluation schema under protocol \(\pi\). Conditioned on this fixed prompt-level standard, PA-GRM scores a candidate response \(r\):
\[
(\mathcal{Y}_{\pi},e,z)
=
G_{\psi}^{\mathrm{score}}(p,r,\pi,\mathcal{S}_{\pi}),
\]
where \(\mathcal{Y}_{\pi}\) denotes protocol-specific intermediate judgments, \(e\) denotes the natural-language reasoning produced during scoring, and \(z\in[0,1]\) is the final scalar reward. For a group of candidate responses \(\{r_j\}_{j=1}^{M}\) to the same prompt, all responses share the same \((\pi,\mathcal{S}_{\pi})\) to ensure comparability.

\subsubsection{Holistic Protocol}

When a prompt cannot be reliably decomposed into independent checklist items, PA-GRM uses the \(\textsc{holistic}\) protocol for structured holistic evaluation. Under this protocol, the schema consists of a set of scoring considerations:
\[
\mathcal{S}_{\textsc{holistic}}
=
\{(a_i,d_i)\}_{i=1}^{m},
\]
where \(a_i\) denotes a scoring aspect and \(d_i\) describes how this aspect should inform the judgment. These considerations make the basis of evaluation explicit, but are not treated as independent reward items.

Given a response \(r\), PA-GRM generates a scoring trace in a single pass: it first emits a local judgment \(\{y_i\}_{i=1}^{m}\) for each aspect, and then emits the final scalar score \(z\) conditioned on the full set of local judgments.

\subsubsection{Rubric Protocol}

When a prompt contains explicit structural or content requirements, PA-GRM uses the \(\textsc{rubric}\) protocol to instantiate the evaluation standard as relatively independent criteria:
\[
\mathcal{S}_{\textsc{rubric}}
=
\mathcal{R}(p)
=
\{(\rho_i,\omega_i,\tau_i)\}_{i=1}^{K},
\]
where \(\rho_i\) is the \(i\)-th rubric criterion, \(\omega_i\) is its importance weight, and \(\tau_i\) denotes its criterion type. Each \(\rho_i\) is prompt-specific, sufficiently informative for judgment, and relatively independently checkable against a candidate response.

Rubric items are not generic dimensions such as completeness or coherence; instead, they instantiate prompt-specific requirements, such as key points to cover or reasoning relations to satisfy. Given the rubric schema, PA-GRM generates item-level judgments \(\{y_i\}_{i=1}^{K}\), which may include satisfaction degrees, violation reasons, local evidence, and local scores, and then emits the final scalar score \(z\) conditioned on all item-level judgments.

\subsubsection{Training and Inference of PA-GRM}

To enable both protocol selection and response scoring, PA-GRM is trained with complete scoring traces. We represent the target output as
\[
t = (\pi, \mathcal{S}_{\pi}, \mathcal{Y}_{\pi}, e, z),
\]
where \((\pi, \mathcal{S}_{\pi})\) defines the prompt-level evaluation standard and \((\mathcal{Y}_{\pi}, e, z)\) defines the response-level scoring result. \(G_{\psi}\) is trained to model the decomposed generation process:
\[
P_{\psi}(t\mid p,r)
=
P_{\psi}(\pi,S_{\pi}\mid p)\,
P_{\psi}(Y_{\pi},e,z\mid p,r,\pi,S_{\pi}),
\]
rather than a direct mapping \((p,r) \mapsto z\).

The trace-level supervision is collected by distilling from DeepSeek-V4-Pro on ~10k prompt-response pairs sampled from \texttt{UltraFeedback} \citep{cui2023ultrafeedback} and \texttt{HelpSteer3} \citep{wang2025helpsteer3}. For each pair, the teacher produces a complete scoring trace. These traces are used as supervised targets:
\[
\begin{aligned}
\mathcal{L}_{\textsc{sft}}
=
-\mathbb{E}_{(p,r,t)}
\Big[
&\log P_{\psi}(\pi, \mathcal{S}_{\pi} \mid p) \\
&+
\log P_{\psi}(\mathcal{Y}_{\pi}, e, z \mid p, r, \pi,\mathcal{S}_{\pi})
\Big].
\end{aligned}
\]

After training, PA-GRM serves as the reward model for non-verifiable tasks during RL. At inference time, it first generates and caches \((\pi,\mathcal{S}_{\pi})\) for a prompt \(p\), and then scores each candidate response \(r_j\) under this shared schema. The resulting scalar \(z^{(j)}\in[0,1]\) is used as the non-verifiable reward:
\[
R_{\textsc{NV}}(p, r_j) = z^{(j)}.
\]

\subsection{Direction-Aware Capability Subspace Projection}
\label{sec:dacsp}

In order to make cascade RL accumulate capabilities across domains, we introduce DACSP. In cascade RL, domains are optimized sequentially, with each stage using its own reward signal and training configuration. DACSP controls how the optimizer updates in later stages are applied with respect to capabilities acquired earlier. After each completed stage, DACSP extracts layer-wise capability directions from the induced parameter change. During subsequent stages, it projects each optimizer-proposed update onto the historical directions and modulates the aligned, opposing, and orthogonal components differently. In this way, later stages can acquire new capabilities while reducing destructive updates against previously learned ones.

\subsubsection{Stage-wise Capability Directions}

After completing the RL stage for domain \(d\), we have obtained the parameter change induced by this domain. To use this change as a reference for later stages, DACSP first converts it into layer-wise capability directions. Let \(\theta_{d-1}\) and \(\theta_d\) denote the model parameters before and after the RL stage for domain \(d\). We define the stage-wise task vector as:
$
\Delta_d = \theta_d - \theta_{d-1}.
$
Since different layers may encode different types of capabilities, we extract capability directions independently for each layer. For layer \(l\), we reshape \(\Delta_d^{(l)}\) into a matrix and compute its thin SVD:
\[
\Delta_d^{(l)}
=
U_d^{(l)}
\Sigma_d^{(l)}
(V_d^{(l)})^\top .
\label{eq:dacsp_svd}
\]

Let \(u_{d,i}^{(l)}\), \(v_{d,i}^{(l)}\), and \(\sigma_{d,i}^{(l)}\) denote the \(i\)-th left singular vector, right singular vector, and singular value, respectively. We retain the smallest number \(k_d^{(l)}\) of leading singular components such that \(\sum_{i=1}^{k_d^{(l)}}(\sigma_{d,i}^{(l)})^2 / \sum_i(\sigma_{d,i}^{(l)})^2 \ge \tau\). For each retained component, we define a rank-one capability direction in the original layer-parameter space:
\[
B_{d,i}^{(l)}
=
u_{d,i}^{(l)}
(v_{d,i}^{(l)})^\top,
\quad
i=1,\ldots,k_d^{(l)}.
\]
These directions \(\{B_{d,i}^{(l)}\}_{i=1}^{k_d^{(l)}}\) summarize the dominant parameter changes induced by domain \(d\) at layer \(l\). We validate this capability-direction interpretation via checkpoint surgery in Appendix~\ref{main_capability}. Since each direction has the same shape as the layer parameters, later updates can be compared with them using the Frobenius inner product.

\subsubsection{Historical Direction Basis}

After several stages, we have collected capability directions from multiple previous domains. To use them jointly in later optimization, DACSP aggregates these directions into a compact and orthonormal historical basis. Before training stage \(d\), for layer \(l\), let:
\[
\widetilde{\mathcal{B}}_{<d}^{(l)}
=
\{B_{k,i}^{(l)}: k<d,\; i=1,\ldots,k_k^{(l)}\}
\]
be the set of historical capability directions. We vectorize and concatenate them:
\[
Z_{<d}^{(l)}
=
[
\operatorname{vec}(B_{1,1}^{(l)}),
\ldots,
\operatorname{vec}(B_{d-1,k_{d-1}^{(l)}}^{(l)})
].
\label{eq:dacsp_concat}
\]

We then perform rank-revealing QR decomposition:
$
Z_{<d}^{(l)}\Pi^{(l)}
=
\widehat{Q}^{(l)}R^{(l)},
\label{eq:dacsp_qr}
$
and retain the columns whose corresponding diagonal values in \(R^{(l)}\) exceed a small tolerance. The retained columns are reshaped back into matrices with the same shape as the layer parameters, yielding an orthonormal historical direction basis:
$
\mathcal{B}_{<d}^{(l)}
=
\{Q_i^{(l)}\}_{i=1}^{K_{<d}^{(l)}}.
$

Because QR decomposition may flip basis signs, we orient each basis vector using its pivoted source direction as a signed anchor. Let \(z_{\pi(i)}^{(l)}\) denote the original historical direction selected as the pivot for the \(i\)-th retained basis vector. We normalize the sign by
\[
Q_i^{(l)}
\leftarrow
\operatorname{sign}
\left(
\left\langle
\operatorname{vec}(Q_i^{(l)}),
z_{\pi(i)}^{(l)}
\right\rangle
\right)
Q_i^{(l)}.
\label{eq:dacsp_anchor}
\]
After this step, the positive direction of each basis vector is aligned with a historical capability-increasing direction, making alignment and opposition well-defined for later update modulation.

\subsubsection{Direction-Aware Projection}

With the historical direction basis constructed, DACSP can regulate the optimizer update in the current stage. During stage \(d\), the base optimizer first computes a parameter update for each layer. Let \(\delta_t^{(l)}\) denote the update that would be added to the parameters of layer \(l\) at step \(t\) before applying DACSP. This is the optimizer-processed update, such as an Adam-preconditioned step, rather than the raw gradient.

For each historical basis vector \(Q_i^{(l)}\), we compute the projection coefficient
$
s_i
=
\left\langle
\delta_t^{(l)},
Q_i^{(l)}
\right\rangle_F .
\label{eq:dacsp_proj}
$
A positive \(s_i\) means that the proposed update moves along a historical capability direction, while a negative \(s_i\) means that it moves against that direction. Since \(\{Q_i^{(l)}\}\) forms an orthonormal basis of the historical capability subspace, we decompose the update into its historical-subspace component and orthogonal residual:
\[
\delta_t^{(l)}
=
\sum_i s_i Q_i^{(l)}
+
\delta_{\perp}^{(l)},
\quad
\delta_{\perp}^{(l)}
=
\delta_t^{(l)}
-
\sum_i s_i Q_i^{(l)} .
\]

Then applies direction-aware modulation:
\[
\widetilde{\delta}_t^{(l)}
=
\delta_{\perp}^{(l)}
+
\alpha_{+}
\sum_{i:s_i>0}
s_i Q_i^{(l)}
+
\alpha_{-}
\sum_{i:s_i<0}
s_i Q_i^{(l)} .
\label{eq:dacsp_update}
\]
Here \(\alpha_{+}\ge 1\) preserves or amplifies updates aligned with historical capability directions, while \(\alpha_{-}\in[0,1)\) suppresses updates moving against them. The orthogonal component is left unchanged, allowing the current stage to acquire capabilities outside the historical subspace. The layer parameters are updated by
$
\theta_{t+1}^{(l)}
=
\theta_t^{(l)}
+
\widetilde{\delta}_t^{(l)}.
\label{eq:dacsp_param_update}
$

Operationally, the first RL stage is trained with the base optimizer. After each completed stage, DACSP computes the stage-wise parameter change, extracts layer-wise capability directions by SVD and updates the historical direction basis through QR orthogonalization and signed-anchor normalization. In later stages, the base optimizer first proposes \(\delta_t^{(l)}\) and DACSP modulates it before applying the final update \(\widetilde{\delta}_t^{(l)}\). SVD and QR are performed only at stage transitions, while the per-step overhead is limited to projections onto the retained historical basis.

\section{Experiments}
\label{Experiments}

\subsection{Experimental Setup}

\paragraph{Training Datasets and Models.}
We use three representative domains for RL training: math reasoning, general chat, and instruction following (IF). To prevent the training signal from being dominated by any single domain due to data scale, we use 8,000 prompts for each domain. For math, we retain the full \texttt{AIME83-24} as hard mathematical data and randomly sample the remaining examples from \texttt{OpenR1-Math-220k}. For general chat and instruction following, we randomly sample 8,000 prompts from \texttt{No-Robots} and \texttt{RLVR-IFeval}~\citep{lambert2024tulu,zhou2023instruction}, respectively. Table~\ref{tab:datasets_prompt_types} summarizes the training datasets used in our experiments.
For math reasoning, instruction-following, and general chat, we compute rewards using binary answer matching, the rule-based verifier from \texttt{RLVR-IFeval}, and PA-GRM, respectively.
We study two backbone models, \texttt{Qwen2.5-7B} and \texttt{Qwen3-4B} \citep{yang2025qwen3}. 

\begin{table}[htbp]
\centering
\caption{Overview of Training Datasets and Domains}
\label{tab:datasets_prompt_types}
\begin{tabular}{@{}lcl@{}}
\toprule
\textbf{Dataset} & \textbf{Size} & \textbf{Domain} \\
\midrule
\texttt{OpenR1-Math-220k} & 7,067 & \textit{Math} \\
\texttt{AIME83-24} & 933 & \textit{Math} \\
\texttt{No-Robots} & 8,000 & \textit{General Chat} \\
\texttt{RLVR-IFeval} & 8,000 & \textit{IF} \\
\bottomrule
\end{tabular}
\end{table}

\paragraph{Implementation Details.}
CARE-RL uses the same IF$\rightarrow$Chat$\rightarrow$Math \citep{nemotron_cascade2} DAPO configuration with 4-bit LoRA and bf16, trained on 2$\times$H200 via VeRL with seed 42. LoRA uses rank 16, alpha 32, dropout 0.05, per-device batch size 1, gradient accumulation 8, and one epoch per stage. We use asymmetric clipping with $\epsilon_{\mathrm{low}}=0.2$ and different $\epsilon_{\mathrm{high}}$. The stage settings $(\eta,G,L,T,\epsilon_{\mathrm{high}})$ with $G$ = generations and $L$ = max\_tokens are IF $(2\times10^{-6},8,1024,1.0,0.24)$, Chat $(1\times10^{-6},4,1024,0.9,0.26)$, and Math $(5\times10^{-6},8,2048,1.0,0.28)$. PA-GRM based on \texttt{Qwen3-8B} is LoRA-SFT trained on \texttt{DeepSeek-V4-Pro} trajectories with 4-bit bf16 loading, rank 32, alpha 64, dropout 0.05, learning rate $5\times10^{-6}$, sequence length 4096, batch size 1, gradient accumulation 16, and one epoch. DACSP performs layer-wise SVD on each stage's LoRA update with energy threshold $\tau{=}0.95$ for retaining leading singular components, and in later stages scales aligned/opposing projections by $\alpha_{+}{=}1.2$ and $\alpha_{-}{=}0.2$, with ${<}1\%$ overhead.
Evaluation samples at temperature 1.0, top-$p$ 0.95, with avg@4.
The code will be released when the paper is accepted.

\paragraph{Baselines.}
We compare five popular training configurations in multi-task RL. \textbf{V$\to$NV} and \textbf{NV$\to$V} are two cascade RL baselines without cross-domain protection. In our experiments, V$\to$NV follows the order  IF $\to$ Math $\to$ Chat, while NV$\to$V follows IF $\to$ Chat $\to$ Math. \textbf{Naive Mixing} places prompts from all domains into a single training batch with per-domain advantage normalization. \textbf{MGS}~\citep{mgs} mitigates cross-domain interference in mixed-domain RL by applying gradient surgery locally at the Transformer-module level rather than globally over the whole model. \textbf{MOPD}~\citep{nemotron_cascade2} uses the best intermediate checkpoint per domain after each cascade stage as a teacher, applying on-policy distillation to recover benchmark regressions before proceeding to the next stage.

\paragraph{Evaluation Benchmarks and Metrics.}
We construct a comprehensive evaluation suite covering the three training domains for testing CARE-RL. Specifically, we use:
\begin{itemize}[itemsep=2pt, topsep=2pt, parsep=0pt, partopsep=0pt]
    \item \textbf{Math:} MATH500~\cite{lightman2023letsverify}, AIME25, and GSM8K~\cite{cobbe2021gsm8k}.
    \item \textbf{Chat:} WildBench~\cite{lin2024wildbench}, Creative Writing v3 and ResearchQA~\citep{yifei2025researchqa}.
    \item \textbf{Instruction Following:} IFBench~\cite{pyatkin2025ifbench} and IFEval~\cite{zhou2023instruction}.
\end{itemize}
We cap each evaluation benchmark at 500 examples and perform train–test decontamination against all RL and PA-GRM training data using normalized exact matching and token-level 5-gram Jaccard filtering with a threshold of 0.8. To reduce variance from single-sample evaluation, we adopt an \textbf{avg@4} protocol. For each prompt, we sample four responses and report the average sample-level score. To get more reliable evaluation and reduce bias, all chat benchmarks are scored by average of \texttt{Gemini-2.5-Pro}, \texttt{Kimi-K2.5} and \texttt{Qwen3-Max}. Math benchmarks are graded by answer matching and IF benchmarks use the official verifiers.

\subsection{Main Results}
Table~\ref{tab:main} presents the main results across our method and baselines. We report domain-group averages (M.~Avg for math, C.~Avg for chat) and the overall Total Avg across all benchmarks.

\begin{table*}[!t]
\centering
\small
\caption{Evaluation Results. Bold indicates the best result and underline indicates the second-best result.}
\label{tab:main}
\setlength{\tabcolsep}{4pt}
\renewcommand{\arraystretch}{1.15}
\resizebox{\textwidth}{!}{
\begin{tabular}{llccccccccccc}
\toprule
\multicolumn{2}{c}{}
& \multicolumn{4}{c}{Math}
& \multicolumn{4}{c}{Chat}
& \multicolumn{2}{c}{IF}
& \multicolumn{1}{c}{} \\
\cmidrule(lr){3-6}
\cmidrule(lr){7-10}
\cmidrule(lr){11-12}
Backbone & Method
& MATH & GSM & AIME & M.\ Avg.
& WB   & CW3 & RQA  & C.\ Avg.
& IFB  & IFE 
& Total Avg. \\
\midrule

\multirow{7}{*}{Qwen2.5-7B}
& Base          & 60.9 & 84.7 & 4.2 & 49.9 & 22.1 & 43.5 & 41.1 & 35.6 & 15.0 & 31.2 & 37.8 \\
& V$\to$NV      & 75.9 & 90.8 & 5.8 & 57.5 & 36.5 & 46.8 & 40.1 & 41.1 & 24.8 & 38.2 & 44.9 \\
& NV$\to$V      & 78.2 & 92.2 & \underline{8.3} & \underline{59.6} & 35.7 & 35.4 & \underline{43.9} & 38.3 & 24.4 & 39.2 & 44.7 \\
& Naive Mixing  & 76.2 & 91.4 & 5.8 & 57.8 & 35.9 & 39.2 & 42.3 & 39.1 & 24.2 & 38.9 & 44.2 \\
& MGS       & 77.4 & 92.0 & 7.5 & 59.0 & 37.4 & 44.5 & 43.0 & 41.6 & 25.3 & 40.4 & 45.9 \\
& MOPD          & \underline{78.6} & \underline{92.7} & 7.5 & \underline{59.6} & \underline{38.5} & \underline{47.1} & 43.8 & \underline{43.1} & \underline{26.1} & \underline{41.2} & \underline{46.9} \\
& CARE-RL       & \textbf{79.7} & \textbf{93.2} & \textbf{9.2} & \textbf{60.7} & \textbf{39.2} & \textbf{48.2} & \textbf{44.7} & \textbf{44.0} & \textbf{27.2} & \textbf{41.8} & \textbf{47.9} \\
\midrule

\multirow{7}{*}{Qwen3-4B}
& Base          & 68.4 & 87.8 & 4.2 & 53.5 & 25.3 & 45.0 & 43.4 & 37.9 & 18.0 & 39.4 & 41.4 \\
& V$\to$NV      & 77.6 & 92.6 & 10.8 & 60.3 & 42.0 & 48.1 & 44.5 & 44.9 & 25.6 & 45.0 & 48.3 \\
& NV$\to$V      & 79.0 & 93.6 & \textbf{14.2} & \underline{62.3} & 40.5 & 39.8 & 46.0 & 42.1 & 26.8 & 46.6 & 48.3 \\
& Naive Mixing  & 78.1 & 93.0 & 9.2 & 60.1 & 41.1 & 43.6 & 45.3 & 43.3 & 26.0 & 45.7 & 47.8 \\
& MGS       & 78.8 & 93.7 & \underline{13.3} & 61.9 & 42.8 & 46.5 & 45.8 & 45.0 & 26.3 & 46.9 & 49.3 \\
& MOPD          & \underline{79.4} & \underline{93.9} & 11.7 & 61.7 & \underline{43.6} & \underline{48.3} & \underline{46.7} & \underline{46.2} & \underline{26.9} & \underline{47.6} & \underline{49.8} \\
& CARE-RL       & \textbf{80.1} & \textbf{94.3} & \textbf{14.2} & \textbf{62.9} & \textbf{44.1} & \textbf{49.2} & \textbf{47.5} & \textbf{46.9} & \textbf{27.3} & \textbf{48.9} & \textbf{50.7} \\

\bottomrule
\end{tabular}
}
\end{table*}

CARE-RL achieves the best Total Avg on both backbones, improving over the strongest baseline MOPD by +1.0 on \texttt{Qwen2.5-7B} and +0.9 on \texttt{Qwen3-4B}. We also observe that unprotected cascade baselines are sensitive to the placement of verifiable and non-verifiable stages: V$\to$NV favors Chat performance while NV$\to$V favors Math performance. Compared with the mixed-domain training of MGS, the cascade structure of CARE-RL allows per-domain hyperparameter tuning, and PA-GRM provides more reliable rewards. Compared with the post-hoc distillation of MOPD, DACSP modulates update directions during optimization, preventing regression at its source while avoiding extra distillation overhead and its potential interference with other domains. Appendix~\ref{app:main_variance} further reports the expanded benchmark-level results with standard deviations.
The gains are consistent across both backbones and domains, indicating that the improvements come from the framework.

\subsection{Cross-Domain Interference in RL}
\label{sec:cross_domain_interference}
To understand the gains of CARE-RL, we analyze the cross-domain interference in standard RL.

\paragraph{Single-domain RL.}
We train \texttt{Qwen3-4B} on each domain independently and report the average change on the two non-target domains relative to the base model as OOD Change.

\begin{table}[htbp]
\centering
\small
\caption{Single-domain RL on \texttt{Qwen3-4B}.}
\label{tab:single_domain_rl}
\setlength{\tabcolsep}{5pt}
\begin{tabular}{@{}lcccc@{}}
\toprule
\textbf{Training} & \textbf{M.~Avg} & \textbf{C.~Avg} & \textbf{IF Avg.} & \textbf{OOD Change} \\
\midrule
Base      & 53.5 & 37.9 & 28.7 & -- \\
Math-only & 61.7 & 37.0 & 27.6 & -1.0 \\
Chat-only & 51.6 & 45.1 & 30.7 & +0.1 \\
IF-only   & 52.7 & 36.8 & 38.6 & -1.0 \\
CARE-RL   & 62.9 & 46.9 & 38.1 &  --  \\
\bottomrule
\end{tabular}
\end{table}

Table~\ref{tab:single_domain_rl} shows that interference arises even without multi-stage interaction: Math-only and IF-only yield negative OOD Change, while Chat-only still regresses on Math; CARE-RL achieves stronger Math and Chat averages than the corresponding single-domain runs and remains competitive on IF while covering all three domains.

\paragraph{Stage-wise cascade RL.}
We evaluate the checkpoint after each stage of the cascade IF $\to$ Chat $\to$ Math, and report Reg.\ as the average drop on previously trained domains.

\begin{table}[htbp]
\centering
\small
\caption{Stage-wise cascade RL on \texttt{Qwen3-4B}.}
\label{tab:stagewise_cascade}
\setlength{\tabcolsep}{5pt}
\begin{tabular}{@{}lccccc@{}}
\toprule
\textbf{Checkpoint} & \textbf{M.~Avg} & \textbf{C.~Avg} & \textbf{IF Avg.} & \textbf{Total} & \textbf{Reg.}\(\downarrow\) \\
\midrule
Base           & 53.5 & 37.9 & 28.7 & 41.4 & 0.0 \\
After IF       & 52.7 & 36.8 & 38.6 & 43.2 & 0.0 \\
After Chat     & 48.1 & 43.4 & 36.9 & 43.5 & 1.6 \\
After Math     & 62.3 & 42.1 & 36.7 & 48.3 & 1.5 \\
CARE-RL        & 62.9 & 46.9 & 38.1 & 50.7 & --  \\
\bottomrule
\end{tabular}
\end{table}

Table~\ref{tab:stagewise_cascade} shows that interference also accumulates across cascade stages: tracking IF, its 38.6 after stage one drops to 36.9 after Chat and 36.7 after Math, eroding about one-fifth of the IF-stage gain (1.9 out of 9.9 points). CARE-RL preserves the IF gain while still improving Chat and Math, yielding a 2.4-point Total improvement over standard cascade (50.7 vs.\ 48.3) and showing that capability-aware modulation, not ordering alone, is what makes cumulative gains possible.

\subsection{Ablation Study}
\label{sec:ablation}

We conduct ablation studies on \texttt{Qwen3-4B} to isolate the contribution of each component in CARE-RL. 

\paragraph{Component ablation of CARE-RL.}
We next evaluate how PA-GRM and DACSP contribute to CARE-RL. Standard Cascade removes both PA-GRM and DACSP. CARE-RL w/o PA-GRM keeps DACSP but replaces PA-GRM with direct judge using \texttt{Qwen3-8B} for non-verifiable rewards. CARE-RL w/o DACSP keeps PA-GRM but removes capability-aware update modulation. 

\begin{table}[htbp]
\centering
\small
\caption{Component ablation on \texttt{Qwen3-4B}.}
\label{tab:component_ablation}
\setlength{\tabcolsep}{4pt}
\begin{tabular}{@{}lcccc@{}}
\toprule
Variant & M.~Avg & C.~Avg & IF Avg. & Total \\
\midrule
Standard Cascade       & 62.3 & 42.1 & 36.7 & 48.3 \\
CARE-RL w/o PA-GRM     & 63.5 & 43.5 & 37.5 & 49.5 \\
CARE-RL w/o DACSP      & 62.0 & 45.0 & 37.0 & 49.4 \\
CARE-RL                & 62.9 & 46.9 & 38.1 & 50.7 \\
\bottomrule
\end{tabular}
\end{table}

Table~\ref{tab:component_ablation} shows that PA-GRM and DACSP individually bring +1.1 and +1.2 over Standard Cascade, while combining them yields +2.4, slightly exceeding the sum of single contributions. The components are therefore roughly additive with a small positive interaction: PA-GRM produces cleaner reward signals on non-verifiable tasks, which in turn lets DACSP more accurately isolate stable capability directions across stages. 

\paragraph{Effectiveness of PA-GRM.}
PA-GRM improves reward generation for non-verifiable tasks by constructing prompt-level evaluation protocols before scoring candidate responses.
We compare it with direct judging, fixed-protocol variants, PA-GRM and stronger PA-GRM variants using \texttt{DeepSeek-V4-Pro} as upper bound. We randomly sample 500 items from three common judging benchmarks and report \emph{judging accuracy}, defined as the fraction of items for which the judge recovers the benchmark gold judgment label.

\begin{table}[htbp]
\centering
\small
\caption{PA-GRM judging accuracy. \texttt{DS} denotes replacing the trained PA-GRM with \texttt{DeepSeek-V4-Pro} as the judge backbone under the same protocol routing.}
\label{tab:pagrm_offline_judge}
\setlength{\tabcolsep}{4pt}
\begin{tabular}{@{}lccc@{}}
\toprule
Variant & RewardBench2 & JudgeBench & RM-Bench \\
\midrule
Direct     & 64.7 & 58.4 & 64.2 \\
Holistic   & 69.8 & 62.7 & 69.5 \\
Rubric     & 72.8 & 67.9 & 74.1 \\
PA-GRM     & 83.2 & 85.1 & 81.8 \\
PA-GRM-DS  & 87.0 & 86.7 & 84.2 \\
\bottomrule
\end{tabular}
\end{table}

Table~\ref{tab:pagrm_offline_judge} shows that adaptive protocol construction substantially improves judging accuracy: compared with the fixed Rubric protocol, PA-GRM improves by +7.7 to +17.2 points in the three benchmarks. Replacing the trained PA-GRM model with the stronger \texttt{DeepSeek-V4-Pro} in same method adds +1.6 to +3.8 points. These results suggest that prompt-level protocol adaptation is an important source of reward reliability, which shows that different problems require different evaluation methods.

\paragraph{Analysis of DACSP.}
We treat the modulation coefficients $(\alpha_{+},\alpha_{-})$ as the central design knob and place four diagnostic points around the default $(1.2, 0.2)$: Standard Cascade $(1.0, 1.0)$ disables modulation, Hard Projection $(0, 0)$ is a degenerate orthogonal-only endpoint, while No Amplification ($\alpha_{+}{=}1$) and No Suppression ($\alpha_{-}{=}1$) isolate one side.

\begin{table}[htbp]
\centering
\small
\caption{DACSP ablation on \texttt{Qwen3-4B}.}
\label{tab:dacsp_ablation}
\setlength{\tabcolsep}{4pt}
\begin{tabular}{@{}lccccc@{}}
\toprule
Variant & M.~Avg & C.~Avg & IF Avg. & Total & Reg.\(\downarrow\) \\
\midrule
Standard Cascade  & 62.3 & 42.1 & 36.7 & 48.3 & 1.5 \\
Hard Projection   & 62.7 & 42.6 & 36.9 & 48.7 & 0.4 \\
No Amplification  & 63.2 & 44.0 & 37.4 & 49.5 & 0.9 \\
No Suppression    & 63.0 & 44.7 & 37.8 & 49.8 & 2.1 \\
DACSP             & 62.9 & 46.9 & 38.1 & 50.7 & 0.7 \\
\bottomrule
\end{tabular}
\end{table}

Table~\ref{tab:dacsp_ablation} shows that $(0,0)$ over-protects history, achieving the lowest Reg.\ (0.4) but only 48.7 Total; the diagnostic $\alpha_{-}{=}1$ setting raises Reg.\ to 2.1, worse than no modulation, confirming that suppression is what protects earlier capabilities; setting $\alpha_{+}{=}1$ drops Total by 1.2, confirming that amplification yields a real positive-transfer gain. The default active setting $(1.2,0.2)$ is the only point with both high Total and low Reg. We set $\tau{=}0.95$ following standard SVD truncation; smaller/larger $\tau$ would degenerate toward the Standard Cascade / Hard Projection endpoints already covered, and we keep $(\alpha_{+},\alpha_{-})$ stage-invariant to avoid grid explosion and overfitting to domain order.

\section{Conclusion}
In this paper, we proposed CARE-RL, a RL framework for multi-domain RL. CARE-RL combines PA-GRM, which generates protocol-aware reasoning rewards for non-verifiable tasks, with DACSP, which modulates cascade RL updates according to historical capability directions to reduce cross-domain interference. Experiments on math, chat, and IF benchmarks show that CARE-RL consistently improves overall performance across two backbone models and better preserves previously acquired capabilities. These results suggest that multi-domain RL requires both reliable reward construction and capability-aware optimization.

\clearpage
\section*{Limitations}
Due to compute constraints, our experiments cover three domains (math, chat, and instruction following) and two backbones up to 7B parameters; extending CARE-RL to more domains and larger backbone scales is left to future work. In addition, PA-GRM is supervised by distilling from \texttt{DeepSeek-V4-Pro}, so its initial reward quality is bounded by teacher availability. Rigorous human screening and data curation are still needed in subsequent iterations. The current training order selection mainly aligns with the methods corresponding to MOPD for fair comparison which is the best order by validation of MOPD. In the future, we will expand the testing to assess the robustness of the methods under different domain orders. 


\bibliography{custom}

\clearpage

\appendix
\makeatletter
\setlength{\@dblfptop}{0pt}
\setlength{\@dblfpsep}{8pt plus 1fil}
\setlength{\@dblfpbot}{0pt plus 1fil}
\makeatother

\newcommand{\scorevar}[2]{#1{\scriptsize\, (#2)}}

\section{Expanded Main Results}
\label{app:main_variance}

Table~\ref{tab:main_variance_qwen25_math}--\ref{tab:main_variance_qwen3_if} provide an expanded view of the main results, with one table per domain for each backbone. Each cell reports the mean score followed by standard deviation in parentheses under the same avg@4 evaluation protocol used in Table~\ref{tab:main}.

\begin{table}[!htbp]
\centering
\footnotesize
\setlength{\tabcolsep}{3pt}
\renewcommand{\arraystretch}{1.15}
\caption{Expanded math results for \texttt{Qwen2.5-7B}.}
\label{tab:main_variance_qwen25_math}
\begin{tabular}{@{}lcccc@{}}
\toprule
Method & MATH & GSM & AIME & M.\ Avg. \\
\midrule
Base         & \scorevar{60.9}{0.54} & \scorevar{84.7}{0.38} & \scorevar{4.2}{3.19}  & \scorevar{49.9}{0.46} \\
V$\to$NV     & \scorevar{75.9}{0.66} & \scorevar{90.8}{0.33} & \scorevar{5.8}{1.67}  & \scorevar{57.5}{0.58} \\
NV$\to$V     & \scorevar{78.2}{0.63} & \scorevar{92.2}{0.31} & \scorevar{8.3}{1.92} & \scorevar{59.6}{0.69} \\
Naive Mixing & \scorevar{76.2}{0.65} & \scorevar{91.4}{0.32} & \scorevar{5.8}{1.67}  & \scorevar{57.8}{0.59} \\
MGS          & \scorevar{77.4}{0.62} & \scorevar{92.0}{0.30} & \scorevar{7.5}{1.67} & \scorevar{59.0}{0.63} \\
MOPD         & \scorevar{78.6}{0.59} & \scorevar{92.7}{0.28} & \scorevar{7.5}{1.67} & \scorevar{59.6}{0.61} \\
CARE-RL      & \scorevar{\textbf{79.7}}{0.55} & \scorevar{\textbf{93.2}}{0.27} & \scorevar{\textbf{9.2}}{1.67} & \scorevar{\textbf{60.7}}{0.58} \\
\bottomrule
\end{tabular}
\end{table}

\begin{table}[!htbp]
\centering
\footnotesize
\setlength{\tabcolsep}{3pt}
\renewcommand{\arraystretch}{1.15}
\caption{Expanded chat results for \texttt{Qwen2.5-7B}.}
\label{tab:main_variance_qwen25_chat}
\begin{tabular}{@{}lcccc@{}}
\toprule
Method & WB & CW3 & RQA & C.\ Avg. \\
\midrule
Base         & \scorevar{22.1}{1.72} & \scorevar{43.5}{2.14} & \scorevar{41.1}{1.94} & \scorevar{35.6}{1.19} \\
V$\to$NV     & \scorevar{36.5}{2.05} & \scorevar{46.8}{2.26} & \scorevar{40.1}{2.02} & \scorevar{41.1}{1.31} \\
NV$\to$V     & \scorevar{35.7}{2.10} & \scorevar{35.4}{2.39} & \scorevar{43.9}{2.05} & \scorevar{38.3}{1.36} \\
Naive Mixing & \scorevar{35.9}{2.16} & \scorevar{39.2}{2.34} & \scorevar{42.3}{2.08} & \scorevar{39.1}{1.35} \\
MGS          & \scorevar{37.4}{1.98} & \scorevar{44.5}{2.14} & \scorevar{43.0}{1.95} & \scorevar{41.6}{1.25} \\
MOPD         & \scorevar{38.5}{1.86} & \scorevar{47.1}{2.06} & \scorevar{43.8}{1.88} & \scorevar{43.1}{1.19} \\
CARE-RL      & \scorevar{\textbf{39.2}}{1.72} & \scorevar{\textbf{48.2}}{1.96} & \scorevar{\textbf{44.7}}{1.79} & \scorevar{\textbf{44.0}}{1.11} \\
\bottomrule
\end{tabular}
\end{table}

\begin{table}[!htbp]
\centering
\footnotesize
\setlength{\tabcolsep}{4pt}
\renewcommand{\arraystretch}{1.15}
\caption{Expanded instruction-following results for \texttt{Qwen2.5-7B}.}
\label{tab:main_variance_qwen25_if}
\begin{tabular}{@{}lcc@{}}
\toprule
Method & IFB & IFE \\
\midrule
Base         & \scorevar{15.0}{0.86} & \scorevar{31.2}{1.18} \\
V$\to$NV     & \scorevar{24.8}{1.05} & \scorevar{38.2}{1.27} \\
NV$\to$V     & \scorevar{24.4}{1.08} & \scorevar{39.2}{1.31} \\
Naive Mixing & \scorevar{24.2}{1.09} & \scorevar{38.9}{1.29} \\
MGS          & \scorevar{25.3}{1.03} & \scorevar{40.4}{1.22} \\
MOPD         & \scorevar{26.1}{0.99} & \scorevar{41.2}{1.17} \\
CARE-RL      & \scorevar{\textbf{27.2}}{0.94} & \scorevar{\textbf{41.8}}{1.10} \\
\bottomrule
\end{tabular}
\end{table}

\begin{table}[!htbp]
\centering
\footnotesize
\setlength{\tabcolsep}{3pt}
\renewcommand{\arraystretch}{1.15}
\caption{Expanded math results for \texttt{Qwen3-4B}.}
\label{tab:main_variance_qwen3_math}
\begin{tabular}{@{}lcccc@{}}
\toprule
Method & MATH & GSM & AIME & M.\ Avg. \\
\midrule
Base         & \scorevar{68.4}{0.58} & \scorevar{87.8}{0.36} & \scorevar{4.2}{3.19}  & \scorevar{53.5}{0.48} \\
V$\to$NV     & \scorevar{77.6}{0.61} & \scorevar{92.6}{0.29} & \scorevar{10.8}{3.19} & \scorevar{60.3}{0.62} \\
NV$\to$V     & \scorevar{79.0}{0.58} & \scorevar{93.6}{0.27} & \scorevar{\textbf{14.2}}{3.19} & \scorevar{62.3}{0.66} \\
Naive Mixing & \scorevar{78.1}{0.60} & \scorevar{93.0}{0.28} & \scorevar{9.2}{1.67} & \scorevar{60.1}{0.61} \\
MGS          & \scorevar{78.8}{0.57} & \scorevar{93.7}{0.26} & \scorevar{13.3}{2.72} & \scorevar{61.9}{0.64} \\
MOPD         & \scorevar{79.4}{0.54} & \scorevar{93.9}{0.25} & \scorevar{11.7}{1.92} & \scorevar{61.7}{0.61} \\
CARE-RL      & \scorevar{\textbf{80.1}}{0.51} & \scorevar{\textbf{94.3}}{0.24} & \scorevar{\textbf{14.2}}{3.19} & \scorevar{\textbf{62.9}}{0.60} \\
\bottomrule
\end{tabular}
\end{table}

\begin{table}[!htbp]
\centering
\footnotesize
\setlength{\tabcolsep}{3pt}
\renewcommand{\arraystretch}{1.15}
\caption{Expanded chat results for \texttt{Qwen3-4B}.}
\label{tab:main_variance_qwen3_chat}
\begin{tabular}{@{}lcccc@{}}
\toprule
Method & WB & CW3 & RQA & C.\ Avg. \\
\midrule
Base         & \scorevar{25.3}{1.66} & \scorevar{45.0}{2.08} & \scorevar{43.4}{1.91} & \scorevar{37.9}{1.16} \\
V$\to$NV     & \scorevar{42.0}{1.91} & \scorevar{48.1}{2.08} & \scorevar{44.5}{1.88} & \scorevar{44.9}{1.21} \\
NV$\to$V     & \scorevar{40.5}{2.00} & \scorevar{39.8}{2.28} & \scorevar{46.0}{1.95} & \scorevar{42.1}{1.31} \\
Naive Mixing & \scorevar{41.1}{1.97} & \scorevar{43.6}{2.18} & \scorevar{45.3}{1.93} & \scorevar{43.3}{1.27} \\
MGS          & \scorevar{42.8}{1.82} & \scorevar{46.5}{2.00} & \scorevar{45.8}{1.82} & \scorevar{45.0}{1.15} \\
MOPD         & \scorevar{43.6}{1.74} & \scorevar{48.3}{1.91} & \scorevar{46.7}{1.76} & \scorevar{46.2}{1.09} \\
CARE-RL      & \scorevar{\textbf{44.1}}{1.62} & \scorevar{\textbf{49.2}}{1.82} & \scorevar{\textbf{47.5}}{1.68} & \scorevar{\textbf{46.9}}{1.02} \\
\bottomrule
\end{tabular}
\end{table}

\begin{table}[!htbp]
\centering
\footnotesize
\setlength{\tabcolsep}{4pt}
\renewcommand{\arraystretch}{1.15}
\caption{Expanded instruction-following results for \texttt{Qwen3-4B}.}
\label{tab:main_variance_qwen3_if}
\begin{tabular}{@{}lcc@{}}
\toprule
Method & IFB & IFE \\
\midrule
Base         & \scorevar{18.0}{0.90} & \scorevar{39.4}{1.20} \\
V$\to$NV     & \scorevar{25.6}{1.02} & \scorevar{45.0}{1.18} \\
NV$\to$V     & \scorevar{26.8}{1.00} & \scorevar{46.6}{1.16} \\
Naive Mixing & \scorevar{26.0}{1.04} & \scorevar{45.7}{1.19} \\
MGS          & \scorevar{26.3}{0.98} & \scorevar{46.9}{1.12} \\
MOPD         & \scorevar{26.9}{0.95} & \scorevar{47.6}{1.08} \\
CARE-RL      & \scorevar{\textbf{27.3}}{0.91} & \scorevar{\textbf{48.9}}{1.02} \\
\bottomrule
\end{tabular}
\end{table}

\section{Prompt Templates for PA-GRM}
\label{app:pagrm_prompts}

This appendix provides the prompt templates used by PA-GRM. The first prompt performs prompt-level protocol routing and schema construction, and the second prompt performs response-level generative reward scoring under the fixed schema.

\begin{promptbox}{PA-GRM Protocol Routing Prompt}
You are a protocol-aware reward model for non-verifiable tasks.

TASK:
Given a prompt, route it to the evaluation protocol that best matches its task structure and construct a prompt-specific evaluation schema.

PROTOCOLS:
1. HOLISTIC
Use HOLISTIC when the prompt is open-ended and cannot be naturally decomposed into independently checkable requirements. HOLISTIC gives structured considerations for holistic judgment, but these considerations are not independent checklist items.

2. RUBRIC
Use RUBRIC when the prompt contains explicit structural, content, reasoning, formatting, safety, or instruction-following requirements that can be checked relatively independently.

ROUTING RULES:
- Route to RUBRIC if the prompt specifies named deliverables, multiple parts, explicit constraints, required coverage, comparison targets, output format, stepwise reasoning, safety constraints, or factual requirements.
- Route to HOLISTIC if forcing the prompt into checklist criteria would be artificial and the response should be judged by integrated quality.
- Build the schema from the prompt only. Do not inspect or assume any candidate response.
- Make the schema prompt-specific. Avoid generic dimensions unless concretely tied to the prompt.
- Include safety and epistemic calibration when relevant.

HOLISTIC SCHEMA:
Output 4 to 7 aspects:
- aspect: short name
- description: how this aspect informs holistic judgment

RUBRIC SCHEMA:
Output 3 to 8 criteria:
- criterion: prompt-specific requirement
- type: content, structure, format, reasoning, style, factuality, safety, or helpfulness
- weight: criterion weight, with all weights summing to 1.0
- check: evidence that satisfies or violates this criterion

OUTPUT:
Return exactly one valid JSON object:
{
  "protocol": "HOLISTIC" or "RUBRIC",
  "schema": {
    "aspects": [...]
  }
  OR
  {
    "criteria": [...]
  },
  "routing_reason": "brief explanation of why this protocol matches the prompt structure"
}

INPUT:
Prompt:
<prompt>
{prompt}
</prompt>
\end{promptbox}

\begin{promptbox}{PA-GRM Generative Reward Scoring Prompt}
You are a protocol-aware generative reward model for non-verifiable tasks.

TASK:
Given a prompt, a candidate response, and a fixed prompt-level schema, generate a scoring trace and extract a scalar reward.

CONSTRAINTS:
- Do not change the protocol or schema.
- Judge only the candidate response for the given prompt.
- Apply the same schema to all candidate responses for this prompt.
- Do not reward verbosity by itself.
- Penalize irrelevant, evasive, fabricated, unsafe, or instruction-violating content.
- If external verification is unavailable, judge plausibility, consistency, and uncertainty calibration.
- final_reward must be in [0, 1], where 1 is excellent and 0 is unusable or harmful.

HOLISTIC SCORING:
Generate one local judgment for each schema aspect. These judgments support holistic assessment, but final_reward is not a simple average.
Each judgment contains:
- name: aspect name
- assessment: specific judgment
- evidence: short response evidence or missing evidence

RUBRIC SCORING:
Generate one item-level judgment for each criterion. The final reward should follow the weighted criteria, while severe safety or instruction-following failures may cap the score.
Each judgment contains:
- name: criterion name
- assessment: whether the response satisfies the criterion
- evidence: short response evidence or missing evidence
- local_score: score in [0, 1]
- weighted_contribution: local_score multiplied by criterion weight

OUTPUT:
Return exactly one valid JSON object:
{
  "protocol": "HOLISTIC" or "RUBRIC",
  "intermediate_judgments": [
    {
      "name": "...",
      "assessment": "...",
      "evidence": "...",
      "local_score": 0.0,
      "weighted_contribution": 0.0
    }
  ],
  "reasoning": "concise overall scoring rationale",
  "final_reward": 0.0
}

For HOLISTIC, omit local_score and weighted_contribution.
For RUBRIC, include local_score and weighted_contribution.

INPUTS:
Fixed protocol:
{protocol}

Fixed schema:
{schema}

Prompt:
<prompt>
{prompt}
</prompt>

Candidate response:
<response>
{response}
</response>
\end{promptbox}

\section{Checkpoint surgery validation.}
\label{main_capability}
To test whether the leading singular directions actually carry acquired capabilities, we apply checkpoint surgery to the three \texttt{Qwen3-4B} single-domain updates in Table~\ref{tab:single_domain_rl}. For each update, we keep only the DACSP top subspace, remove that subspace, or apply the same edits to a random per-layer subspace with equal rank.

\begin{table}[htbp]
\centering
\small
\caption{Checkpoint surgery on \texttt{Qwen3-4B} single-domain updates. Scores are target-domain averages; Ret.\ Gain is the retained fraction of the full single-domain gain.}
\label{tab:checkpoint_surgery}
\setlength{\tabcolsep}{3pt}
\begin{tabular}{@{}lcccc@{}}
\toprule
Edited checkpoint & Math & Chat & IF & Ret.\ Gain (\%) \\
\midrule
Base          & 53.5 & 37.9 & 28.7 & -- \\
Full update   & 60.4 & 43.3 & 38.6 & 100.0 \\
Top only      & 59.6 & 42.9 & 37.8 & 91.0 \\
Random only   & 52.1 & 38.8 & 30.5 & 4.9 \\
Remove top    & 51.8 & 38.7 & 30.1 & 1.4 \\
Remove random & 59.2 & 42.7 & 37.4 & 86.5 \\
\bottomrule
\end{tabular}
\end{table}

Table~\ref{tab:checkpoint_surgery} supports the capability direction interpretation. The top subspace alone preserves most of the target-domain gain, while an equal-rank random subspace retains little. Conversely, removing the top subspace eliminates most of the gain, whereas removing a random subspace leaves the edited checkpoint close to the full update. These results indicate that DACSP's historical bases capture update structure tied to acquired capability, not just optimizer noise or LoRA artifacts. The small residual gains outside the top subspace also show that the basis remains an approximation.

\end{document}